\documentclass[sigconf]{acmart}

\usepackage{amsmath}
\usepackage{multirow}
\usepackage{makecell}
\usepackage{graphicx}
\usepackage{dsfont}
\usepackage{color, colortbl}
\usepackage{pifont}
\usepackage{hhline}
\usepackage{amsmath}
\usepackage{subfigure}
\usepackage{enumerate}
\usepackage{bm}
\usepackage{multirow}
\usepackage{stackengine}

\definecolor{fluorescentorange}{rgb}{1.0, 0.75, 0.0}
\definecolor{etonblue}{rgb}{0.59, 0.78, 0.64}
\definecolor{flamingopink}{rgb}{0.99, 0.56, 0.67}
\definecolor{bleudefrance}{rgb}{0.19, 0.55, 0.91}

\definecolor{my_red}{RGB}{229, 116, 111}
\definecolor{my_blue}{RGB}{71, 108, 218}

\DeclareMathOperator*{\argmin}{arg\,min}

\AtBeginDocument{%
  \providecommand\BibTeX{{%
    \normalfont B\kern-0.5em{\scshape i\kern-0.25em b}\kern-0.8em\TeX}}}

\setcopyright{iw3c2w3}
\copyrightyear{2021}
\acmYear{2021}
\setcopyright{iw3c2w3}
\acmConference[WWW '21]{Proceedings of the Web Conference 2021}{April 19--23, 2021}{Ljubljana, Slovenia}
\acmBooktitle{Proceedings of the Web Conference 2021 (WWW '21), April 19--23, 2021, Ljubljana, Slovenia}
\acmPrice{}
\acmDOI{10.1145/3442381.3449981}
\acmISBN{978-1-4503-8312-7/21/04}

\begin{document}

\title{Curriculum CycleGAN for Textual Sentiment Domain Adaptation with Multiple Sources}

\author{Sicheng Zhao}
\authornote{Corresponding author.}
\authornote{Equal contribution.}
\email{schzhao@gmail.com}
\affiliation{%
  \institution{University of California, Berkeley}
  \country{USA}
}

\author{Yang Xiao}
\authornotemark[2]
\email{17822018007@163.com}
\affiliation{%
  \institution{Nankai University}
  \country{China}
}

\author{Jiang Guo}
\authornotemark[2]
\email{jiang_guo@csail.mit.edu}
\affiliation{%
  \institution{Massachusetts Institute of Technology}
  \country{USA}
}

\author{Xiangyu Yue}
\authornotemark[2]
\email{xyyue@berkeley.edu}
\affiliation{%
  \institution{University of California, Berkeley}
  \country{USA}
}

\author{Jufeng Yang}
\email{yangjufeng@nankai.edu.cn}
\affiliation{%
  \institution{Nankai University}
  \country{China}
}

\author{Ravi Krishna}
\email{ravi.krishna@berkeley.edu}
\affiliation{%
  \institution{University of California, Berkeley}
  \country{USA}
}

\author{Pengfei Xu}
\email{xupengfeipf@didiglobal.com}
\affiliation{%
  \institution{Didi Chuxing}
  \country{China}
}

\author{Kurt Keutzer}
\email{keutzer@berkeley.edu}
\affiliation{%
  \institution{University of California, Berkeley}
  \country{USA}
}

\renewcommand{\shortauthors}{Zhao, et al.}

\begin{abstract}
Sentiment analysis of user-generated reviews or comments on products and services in social networks can help enterprises to analyze the feedback from customers and take corresponding actions for improvement. To mitigate large-scale annotations on the target domain, domain adaptation (DA) provides an alternate solution by learning a transferable model from other labeled source domains. Existing multi-source domain adaptation (MDA) methods either fail to extract some discriminative features in the target domain that are related to sentiment, neglect the correlations of different sources and the distribution difference among different sub-domains even in the same source, or cannot reflect the varying optimal weighting during different training stages. In this paper, we propose a novel instance-level MDA framework, named curriculum cycle-consistent generative adversarial network (C-CycleGAN), to address the above issues. Specifically, C-CycleGAN consists of three components: (1) \textit{pre-trained text encoder} which encodes textual input from different domains into a continuous representation space, (2) \textit{intermediate domain generator} with curriculum instance-level adaptation which bridges the gap across source and target domains, and (3) \textit{task classifier} trained on the intermediate domain for final sentiment classification. C-CycleGAN transfers source samples at instance-level to an intermediate domain that is closer to the target domain with sentiment semantics preserved and without losing discriminative features. Further, our dynamic instance-level weighting mechanisms can assign the optimal weights to different source samples in each training stage. We conduct extensive experiments on three benchmark datasets and achieve substantial gains over state-of-the-art DA approaches. Our source code is released at: \url{https://github.com/WArushrush/Curriculum-CycleGAN}.
\end{abstract}




\maketitle

\section{Introduction}

\begin{figure}[!t]
\centering
\includegraphics[width=1.0\linewidth]{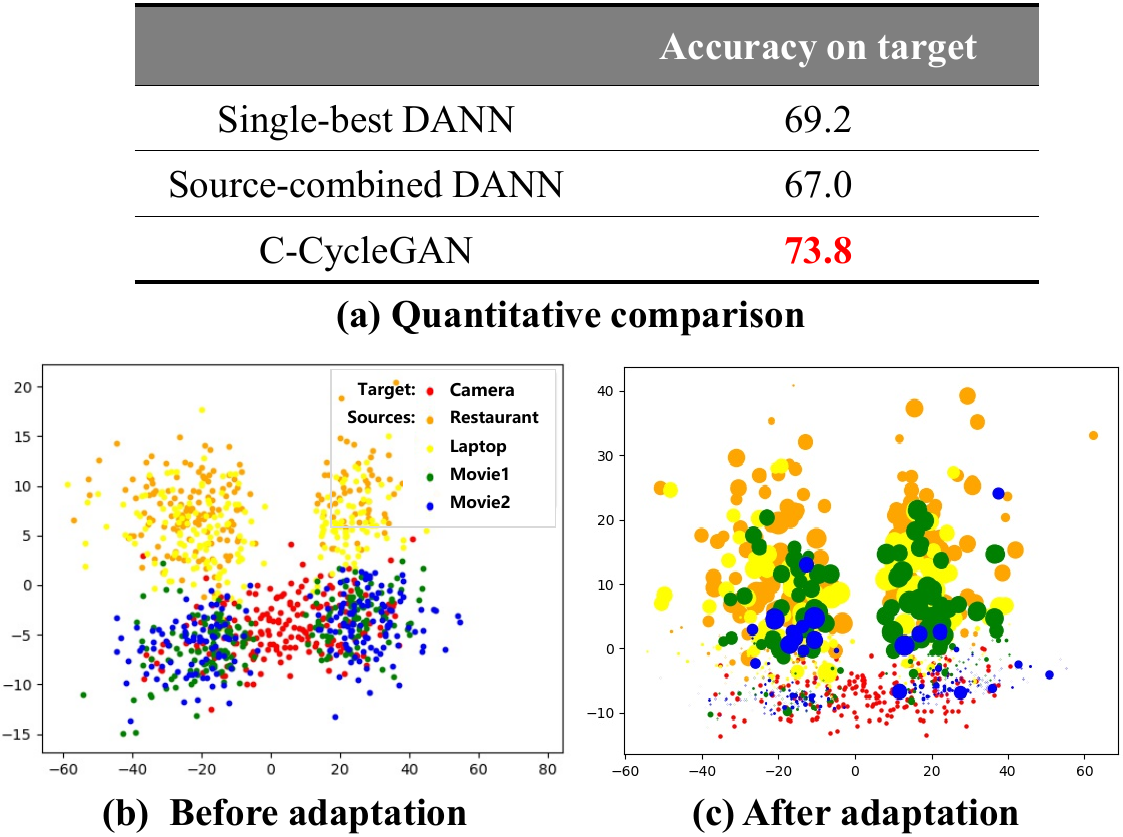}
\caption{An example of \emph{domain shift} in the multi-source scenario on the Reviews-5 dataset~\cite{yu2016learning}, where Camera (red points) is set as the target domain and the rest as source domains. (a) Naively combining multiple sources into one source and directly performing single-source domain adaptation (DANN~\cite{ganin2016domain}) does not guarantee better performance compared to just using the best individual source domain (69.2 vs. 67.0). The proposed C-CycleGAN framework achieves significant performance improvements over the source-trained model baselines (73.8 vs. 69.2). (b) and (c) visualize the representation space before and after adaptation. 
We can see clear domain shift across the sources and the target. After our domain adaptation, the source samples that are closer to the target domain (smaller points) are better aligned to the target domain (larger points indicate smaller sample weights).}
\label{fig:DomainShift}
\end{figure}

The wide popularity of social networks and mobile devices enables human beings to reflect and share their opinions of the products and services they purchase online using text, images, and videos~\cite{zhao2015predicting,zhao2016predicting,deriu2017leveraging,gong2017clustered,luiz2018feature,gong2018sentiment,zhao2020discrete,zhao2020end}. For example, when we plan to buy something, it is of high probability that we take a look at the comments on what others feel about this product. If the negative comments dominate the feedback, we might change our minds to a different brand. Sentiment analysis of user-generated large-scale multimedia data can not only help the customers to select what they want, but also prompt enterprises to improve the quality of their products and services~\cite{zhao2016predicting,chen2019emoji}. Among different multimedia modalities, text, the one focused on in this paper, is the most direct and popular one~\cite{deriu2017leveraging}.

Recent studies~\cite{zhang2018textual, kiritchenko2014sentiment, georgakopoulos2018convolutional,liu2018content,yadav2020sentiment, wang2018sentiment,wang2019aspect,chen2019emoji,biddle2020leveraging} have shown that deep neural networks (DNNs) achieve the state-of-the-art performance on textual sentiment analysis. However, training a DNN to maximize its capacity usually requires large-scale labeled data, which is expensive and time-consuming to obtain. One alternate solution is to train a DNN on a labeled source domain and transfer the DNN to the target domain. However, due to the presence of ``domain shift''~\cite{torralba2011unbiased}, \textit{i.e.} the distribution differences between the source and target domains, direct transfer may result in significant performance degredation~\cite{tzeng2015simultaneous,hoffman2018CyCADA,yue2019domain,yang2020curriculum}. Domain adaptation (DA)~\cite{patel2015visual,sun2015survey,kouw2019review,zhao2020multi,zhao2020review} that aims to minimize the impact of domain shift provides an alternate solution by learning a model on the source domain with high transferability to the target domain.


Current DA methods for textual sentiment analysis mainly focus on the single-source unsupervised setting~\cite{liu2019survey,xi2020domain}, \textit{i.e.} in which there is only one labeled source domain and one unlabeled target domain. While these unsupervised domain adaptation (UDA) methods perform well when the domain gap between the source and target domains is relatively small, they may fail when the domain gap is large or when there are multiple labeled source domains~\cite{guo2018multi,zhao2020multi}, which is a more practical scenario. For example, if we have a target Kitchen domain, which may include reviews on cookbooks, bowls, and electric kettles, and three source domains, books, cookware, and electronics, it is difficult to perfectly align each source and the target.
Naive combination of different sources into one source and direct application of single-source UDA algorithms may lead to suboptimal results, because domain shift also exists across different sources, as shown in Figure~\ref{fig:DomainShift}. Sufficiently exploiting complementary information from different sources can allow for learning a better representation for the target domain, which calls for effective multi-source domain adaptation (MDA) techniques~\cite{sun2015survey,zhao2020multi}.

Recently, some deep MDA approaches have been proposed for textual sentiment classification, most of which are based on adversarial learning, containing a pair of feature extractors and domain classifier (\textit{e.g.} MDAN~\cite{zhao2018adversarial}, MoE~\cite{guo2018multi}). These methods mainly focus on extracting domain-invariant features of different domains, aligning each source and the target separately, or assigning weights to the source samples statically. Although they can obtain domain-invariant features among different domains, there are still some limitations.
First, some discriminative features in the target domain that are related to sentiment might be missing. Since the shared feature extractor mainly aims to extract domain-invariant features by projecting both source samples and target samples to a lower-dimensional space, it may not include all sentiment-related features in the target domain. Second, some existing MDA methods separately align each source and the target and then combine the prediction results with known domain labels, which neglects the correlations of different source domains and different sub-domains even in each source. These methods would naturally fail when the domain labels of labeled source samples are not available. Finally, existing sampling-based methods mainly focus on selecting source samples that are closer to the target by training source selection models to calculate the weight of each sample (\textit{e.g.} MDDA~\cite{zhao2020distilling}, CMSS~\cite{yang2020curriculum}), which cannot reflect the varying optimal weighting during different training stages.

In this paper, we propose a novel instance-level multi-source domain adaptation framework, named curriculum cycle-consistent generative adversarial network (C-CycleGAN), to address the above issues for textual sentiment classification. First, in order to encode all text instances in both source and target domains into a latent continuous representation space with minimal information loss, we introduce text reconstruction to better preserve information.
Second, for the encoded source representations, we generate an intermediate domain to align the mixed source and target domains using a generative adversarial network (GAN) with cycle-consistency. To explore the importance of different source samples in a batch, we assign weights to them at instance-level with novel dynamic model-based and model-free weighting mechanisms. Finally, based on the adapted representations and corresponding source sentiment labels, we train a transferable task classifier. The sentiment loss of the classifier is also backpropagated to the source-to-target generator to preserve the sentiment information before and after generation. Extensive experiments are conducted on three benchmark datasets: Reviews-5~\cite{yu2016learning}, Amazon benchmark~\cite{chen2012marginalized}, and Multilingual Amazon Reviews Corpus~\cite{chen2019multi}. The results show that the proposed C-CycleGAN significantly outperforms the state-of-the-art DA methods for textual sentiment classification.

In summary, the contributions of this paper are threefold:

(1) We propose a novel MDA method, named curriculum cycle-consistent generative adversarial network (C-CycleGAN), to minimize the domain shift between multiple source domains and the target domain. To the best of knowledge, we are the first to generate an intermediate representation domain with cycle-consistency and sentiment consistency for textual sentiment adaptation.

(2) We design novel instance-level model-based and model-free weighting mechanisms, which can update the sample weights dynamically. In this way, our framework does not require domain labels of samples, which allows it to exploit complementary information of all labeled source samples from different domains.

(3) We conduct extensive experiments on three benchmark datasets. As compared to the best baseline, the propsoed C-CycleGAN achieves 1.6\%, 1.2\%, and 13.4\% improvements in average classification accuracy on Reviews-5, Amazon benchmark, and Multilingual Amazon Reviews Corpus, respectively.

\section{Related Work}

\noindent\textbf{Textual Sentiment Analysis.}
Textual sentiment analysis, or opinion mining, aims to assess people's opinions, emotions, and attitudes from text towards entities such as products, services, or organizations~\cite{zhang2018deep}. The wide popularity of social networks such as product reviews, forum discussions, and WeChat, contributes to the rapid development of this task~\cite{zhang2018deep,chen2019emoji}. Traditional sentiment analysis methods mainly focused on designing hand-crafted features~\cite{pang2008opinion,mohammad2013nrc}, which are fed into standard classifiers, such as SVM. Recent efforts on sentiment analysis are mainly based on DNNs~\cite{zhang2018deep,wang2018sentiment}, which have shown great success in many natural language processing tasks. Some typical deep models that have been applied to sentiment analysis include Recursive Auto Encoder~\cite{socher2011semi,dong2014adaptive,qian2015learning}, Recursive Neural Tensor Network~\cite{socher2013recursive}, Recurrent Neural Network (RNN)~\cite{tang2015document}, Long short-term memory (LSTM)~\cite{hochreiter1997long}, Tree-LSTMs~\cite{tai2015improved}, RNN Encoder–Decoder~\cite{cho2014learning}, and BERT~\cite{devlin2019bert}. The above supervised learning methods usually require a large volume of labeled data for training~\cite{liu2019survey,chen2019emoji}. However, high-quality sentiment labels are often labor- and time-consuming to obtain. In this paper, we employ a Bi-LSTM~\cite{hochreiter1997long} as encoder and a multi-layer perceptron as classifier for the sentiment classification adaptation task.


\textbf{Single-source UDA.} Recent single-source UDA (SUDA) methods mainly employ deep learning architectures with two conjoined streams~\cite{zhuo2017deep,zhao2018emotiongan}. One is trained on the labeled source data with a traditional task loss, such as cross-entropy loss for classification. The other aims to align the source and target domains to deal with the domain shift problem with different alignment losses, such as discrepancy loss, adversarial loss, self-supervision loss, \emph{etc}. Discrepancy-based methods employ some distance measurements to explicitly minimize the discrepancy between the source and target domains on specific activation layers, such as maximum mean discrepancies~\cite{long2015learning,wang2018multi,xi2020domain}, correlation alignment~\cite{sun2016return,sun2017correlation,zhuo2017deep}, and contrastive domain discrepancy~\cite{kang2019contrastive}. Adversarial discriminative models usually employ a domain discriminator to adversarially align the extracted features between the source and target domains by making them indistinguishable~\cite{ganin2016domain,tzeng2017adversarial,chen2017no,shen2017wasserstein,tsai2018learning,huang2018domain,kumar2019adversarial,wu2020unsupervised}. Besides the domain discriminator, adversarial generative models also include a generative component to generate fake source or target data typically based on GAN~\cite{goodfellow2014generative} and its variants, such as CoGAN~\cite{liu2016coupled}, SimGAN~\cite{shrivastava2017learning}, and CycleGAN~\cite{zhu2017unpaired,zhao2019cycleemotiongan, hoffman2018CyCADA}. Self-supervision based methods incorporate auxiliary self-supervised learning tasks into the original task network to bring the source and target domains closer. The commonly used self-supervision tasks include reconstruction~\cite{ghifary2015domain,ghifary2016deep,chen2020fido}, image rotation prediction~\cite{sun2019unsupervised,xu2019self}, jigsaw prediction~\cite{carlucci2019domain}, and masking~\cite{vu2020effective}. Although these methods achieve promising results for SUDA tasks, they suffer from significant performance decay when directly applied to MDA task.

\textbf{Multi-source Domain Adaptation.} Based on some theoretical analysis~\cite{ben2010theory,hoffman2018algorithms}, multi-source domain adaptation (MDA) aims to better deal with the scenario where training data are collected from multiple sources~\cite{sun2015survey,zhao2019multi}. The early shallow MDA methods mainly include two categories~\cite{sun2015survey}: feature representation approaches~\cite{sun2011two,duan2012exploiting,chattopadhyay2012multisource,duan2012domain} and combination of pre-learned classifiers \cite{xu2012multi,sun2013bayesian}. Some special MDA cases are considered in recent shallow methods, such as incomplete MDA~\cite{ding2018incomplete} and target shift~\cite{redko2019optimal}.

Recently, some representative deep learning based MDA methods are proposed, such as multisource domain adversarial network (MDAN)~\cite{zhao2018adversarial}, deep cocktail network (DCTN)~\cite{xu2018deep}, Mixture of Experts (MoE)~\cite{guo2018multi}, moment matching network (MMN)~\cite{peng2019moment}, multi-source adversarial domain aggregation network (MADAN)~\cite{zhao2019multi}, multi-source distilling domain adaptation (MDDA)~\cite{zhao2020distilling}, and curriculum manager for source selection (CMSS)~\cite{yang2020curriculum}. MDAN, DCTN, MoE, MMN, MADAN, and MDDA all require domain labels of source samples. MDDA and CMSS select source samples that are closer to the target domain with a static weighting mechanism, while the others do not consider the importance of different source samples. The MDA methods for textual sentiment classification, \textit{e.g.} MDAN and MoE, only focus on extracting domain-invariant features, which may lose discriminative features of the target domain that are related to sentiment. Different from these methods, for the source samples, we generate an intermediate domain that is closer to the target domain with cycle-consistency and sentiment consistency. Further, we propose novel dynamic instance-level weighting mechanisms to assign weights to the source samples without the requirement of domain labels.



\section{Proposed Approach}
\label{sec:Approach}

\begin{figure*}
\centering
\includegraphics[width=0.92\linewidth]{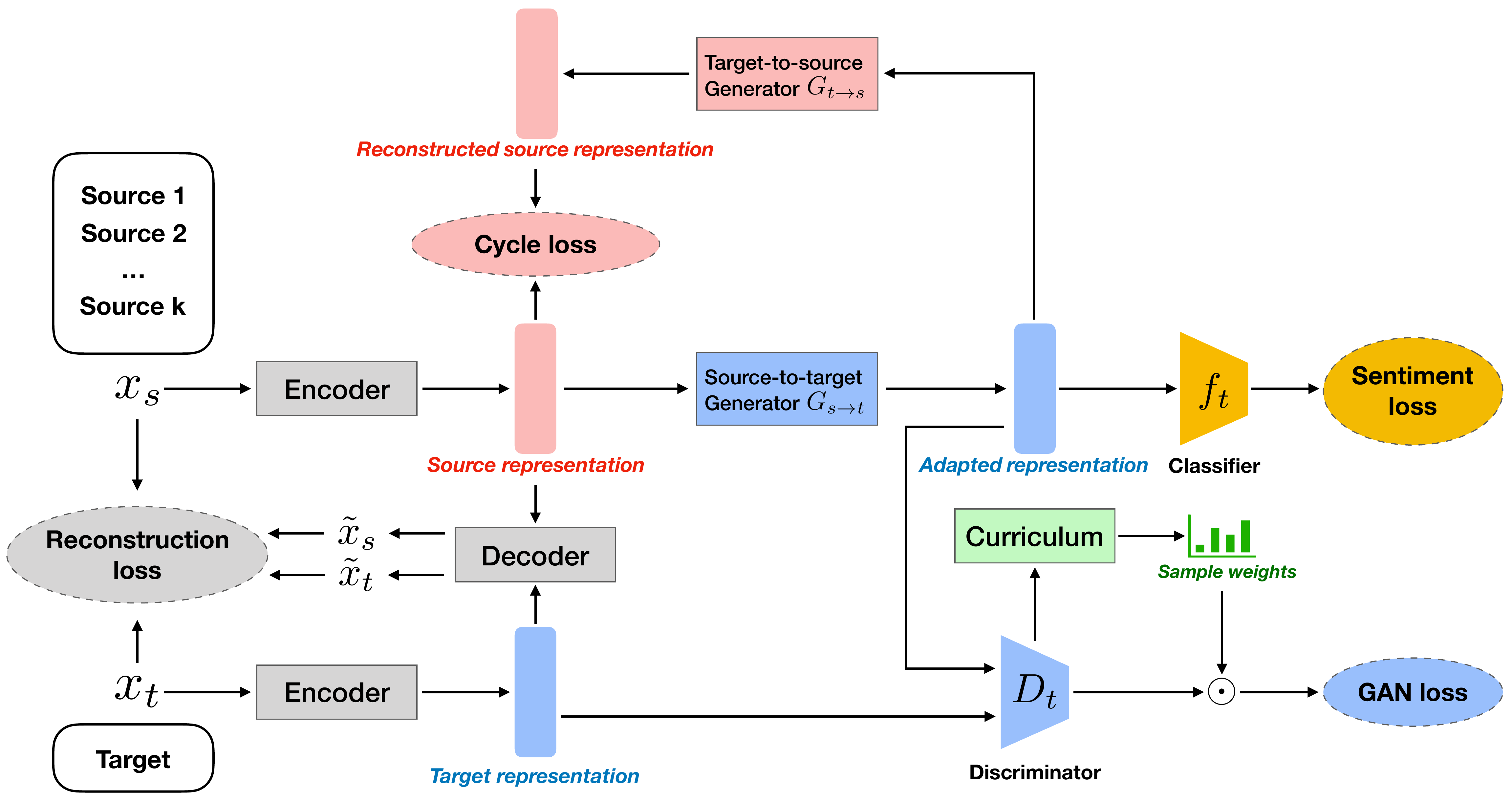}
\caption{Illustration of the proposed C-CycleGAN framework. A text encoder is first pre-trained with a reconstruction loss to encode all text instances from the source and target domains into a latent continuous representation space (\textcolor{gray}{gray}). Then the model is jointly trained using the cycle-consistency loss (\textcolor{flamingopink}{pink}), the curriculum GAN loss (\textcolor{bleudefrance}{\bf blue}), and the sentiment classification loss (\textcolor{fluorescentorange}{yellow}). We depict here the model-free curriculum (\textcolor{etonblue}{green}) for sample weighting.}
\label{fig:model}
\end{figure*}

In this section, we formally define the MDA problem, give an overview of the proposed Curriculum CycleGAN (C-CycleGAN) framework, present each component of C-CycleGAN in detail, and finally introduce the joint learning process.

\subsection{Problem Definition}
We consider the multi-source unsupervised domain adaptation setup for textual sentiment classification, under the \emph{covariate shift} assumption~\cite{patel2015visual}.
Assuming access to $k$ source domains with labeled training data, denoted by $\{\mathcal{S}_i\}_{i=1}^{k}$, where each domain $\mathcal{S}_i$ contains a set of examples drawn from a joint distribution $p^{(S_i)}(\text{x}, \text{y})$ on the input space $\mathcal{X}$ and the output space $\mathcal{Y}$, we seek to learn a sentiment classifier $f: \mathcal{X} \rightarrow \mathcal{Y}$ that is transferable to a target domain $\mathcal{T}$, where only unlabeled data is available.

\subsection{Overview}
Our model bridges the domain gap by generating an intermediate domain using CycleGAN~\cite{zhu2017unpaired} trained with a learned curriculum (C-CycleGAN).
As shown in Figure \ref{fig:model}, the proposed framework has three primary components:

\textit{Pre-trained Text Encoder:} Encode texts from source and target domains into a semantic-preserving latent continuous representation space $\mathcal{Z}$. This module is pre-trained using a \emph{seq2seq}-based text autoencoder in an unsupervised fashion.

\textit{Intermediate Domain Generator:} Generate an intermediate domain to align the multiple sources and the target. At the core of this component is a \emph{curriculum cycle-consistent generative adversarial network}, which employs a domain adversarial loss for distributional alignment, and use cycle-consistency to prevent \emph{mode collapse}. To deal with the varied relevance of the mixed-source instances to the target domain at a specific training stage, we learn a curriculum to dynamically assign weights to source samples based on their proximity to the target domain distribution.

\textit{Task Classifier:} Train the sentiment classifier based on the adapted representations in the intermediate domain and corresponding sentiment labels in the source domains.


\subsection{Pre-trained Text Encoder}
\label{ssec:Text Reconstruction}

We use seq2seq-based text reconstruction to pre-train our text encoder, in order to obtain a semantic-preserving latent representation space.
Let $x$ denote a sequence of tokens $w_1, ..., w_L$, where $L$ is the sequence length.
The reconstruction process can be summarized as the following \emph{encoding-decoding} paradigm:
\begin{equation}
    \mathbf{z} = \mathtt{Enc}(x; \bm{\theta});~~~~\Tilde{x} = \mathtt{Dec}(\mathbf{z}, x; \bm{\psi})
\end{equation}
where $\mathbf{z}$ is the text representation.
We use a bidirectional LSTM (Bi-LSTM)~\cite{hochreiter1997long} as the encoder, and obtain the representation $\mathbf{z}$ of an input sequence by concatenating the last states of forward LSTM and backward LSTM.
A unidirectional LSTM then reconstructs $x$ autoregressively conditioned on $\mathbf{z}$.
At each time step of generation, we randomly sample from the ground-truth token and the generated token as input for the next token prediction.
The overall reconstruction loss over both source and target domain data can thus be written as:
\begin{equation}
    \mathcal{L}_{\text{rec}} = \mathbb{E}_{x\sim \mathcal{X}_S \cup \mathcal{X}_T} \Big[-\frac{1}{L}\sum_{t=1}^{L} \log P(\Tilde{x}_{t}|x_{<t}, \tilde{x}_{<t}, \mathbf{z})\Big]
\end{equation}
After pre-training, the encoder will be fixed and the encoded representations will be directly used for the generation of the latent intermediate domain (Section \ref{ssec:Spatially-Adaptive Instance transfer}).

Alternatively, we can directly use publicly available text encoders like BERT \cite{devlin2019bert}, which are designed to be general-purpose and pre-trained in a self-supervised fashion on a mixture of data sources.
In this study, we experiment with BERT, and take the hidden state of the ``[CLS]'' token as the text representation.\footnote{Note that the ``[CLS]'' representation is typically used as a text representation at the fine-tuning stage with supervision from end tasks, whereas we adopt it here as an unsupervised text representation.}

\subsection{Intermediate Domain Generator}
\label{ssec:Spatially-Adaptive Instance transfer}
\begin{figure}[t]
    \centering
    \includegraphics[width=0.95\linewidth]{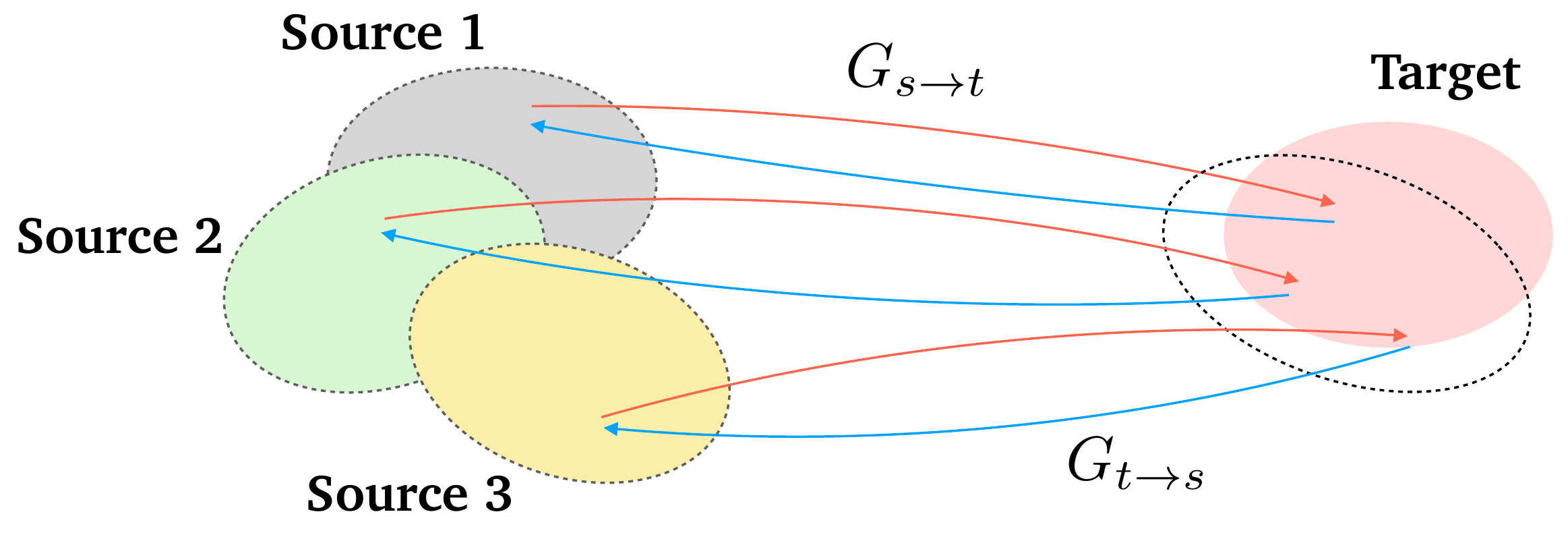}
    \caption{Intermediate domain generation with a CycleGAN.}
    \label{fig:cycle-consistency}
\end{figure}

\textbf{GAN with Cycle-consistency.}
This module generates an intermediate representation domain from the pre-trained representation space $\mathcal{Z}$ to bridge the gap across source and target, as shown in Figure \ref{fig:cycle-consistency}.
For that purpose, we introduce a source-to-target generator $G_{s\rightarrow t}$, and train it to generate target representations that aim to fool an adversarial discriminator $D_t$.
This gives the following GAN loss:
\begin{equation}
\label{eqn:gan-loss-1}
    \mathcal{L}_{\text{gan}}^{s\rightarrow t}=\mathbb{E}_{\mathbf{z}\sim \mathcal{Z}_S}\log [D_t\big(G_{s\rightarrow t}(\mathbf{z})\big)]+\mathbb{E}_{\mathbf{z}\sim \mathcal{Z}_T}\log [1-D_t(\mathbf{z})]\\
\end{equation}

In order to avoid \textit{mode collapse} in the generated intermediate domain and encourage the internal structural information of the original example to be preserved, we follow \cite{hoffman2016fcns} and optimize a \emph{cycle-consistency} loss which is obtained by recontructing the representation of the original example from the intermediate domain representation.
To implement this loss, we introduce a reverse generator from target to source $G_{t\rightarrow s}$, which can be trained using a reverse GAN loss $\mathcal{L}_{\text{adv}}^{t\rightarrow s}$ (this requires an additional discriminator at source side $D_s$). Then, the cycle-consistency loss can be written as:
\begin{equation}
    \begin{split}
    \mathcal{L}_{cyc} &= \mathbb{E}_{\mathbf{z}\sim \mathcal{Z}_S} \parallel G_{t\rightarrow s}\big(G_{s\rightarrow t}(\mathbf{z})\big) - \mathbf{z} \parallel_1\\
                       &+\mathbb{E}_{\mathbf{z}\sim \mathcal{Z}_T} \parallel G_{s\rightarrow t}\big(G_{t\rightarrow s}(\mathbf{z})\big) - \mathbf{z} \parallel_1\\
    \end{split}
\end{equation}


The above loss function treats all source examples in a training batch equally, while ignoring their varied relevance/importance to the target domain distribution due to the multi-source nature.
To cope with this challenge, we explore two instance-level weight assignment mechanisms which operate on each batch: the \emph{model-based curriculum} and the \emph{model-free curriculum}.

\textbf{Model-based Curriculum.} We follow \cite{yang2020curriculum} and use an extra source selection network for calculating the weight distribution over examples in a batch. This network takes the generated representation ($G_{s\rightarrow t}(\mathbf{z})$) as input, and outputs a weight distribution with a Softmax layer.
Denoting $B$ as a batch of encoded examples sampled from the $\mathcal{Z}$: $\{\mathbf{z}_1, \mathbf{z}_2, ..., \mathbf{z}_{|B|}\}$, the sample weights can be computed as:
\begin{equation}
    \mathbf{w} = \mathtt{softmax}\Big(h_t\big(G_{s\rightarrow t}(B)\big)\Big)
\end{equation}
where $h_t$ is the source selection network at target side.
We then obtain the curriculum GAN loss ($\mathcal{L}_{cgan}$) as:
\begin{equation}
    \begin{split}
        \mathcal{L}_{\text{cgan}}^{s\rightarrow t}&=\mathbb{E}_{B\sim \mathcal{Z}_S}\frac{1}{|B|}\sum_{\mathbf{z}\in B} w_z \log [D_t\big(G_{s\rightarrow t}(B)\big)]\\
        &+\mathbb{E}_{\mathbf{z}\sim \mathcal{Z}_T}\log [1-D_t(\mathbf{z})]
    \end{split}
\end{equation}

In the ideal case, if the input batch of the source selection network is extremely close to the \emph{target} distribution, we would expect a uniform weighting.
Therefore, we introduce additional inductive bias for training $h_t$ by minimizing the KL-divergence between the output distribution and a uniform distribution ($\mathcal{U}$) when the input batch is sampled from the real target space:
\begin{equation}
    \mathcal{L}_{uni}^{t} = \mathbb{E}_{\mathbf{z}\sim \mathcal{Z}_T} \mathrm{KL}[h_t(\mathbf{z}) \parallel \mathcal{U}]
\end{equation}

The formulation of $\mathcal{L}_{cgan}^{t\rightarrow s}$ and $\mathcal{L}_{uni}^{s}$ can be adapted in a similar way, using a separate source selection network $h_s$.

\textbf{Model-free Curriculum.} Instead of relying on an extra source selection network, we can also compute sample weights directly from outputs of the domain discriminators ($D_t$), which indeed reflects the proximity of each example to the target domain.
This gives us the following model-free weight assignment mechanism:
\begin{equation}
    \mathbf{w} = \mathtt{softmax}\Big(\log \big[D_t\big(G_{s\rightarrow t}(B)\big)\big]\Big)
\end{equation}
In this way, examples with a higher probability of being classified as \emph{target} will be more emphasized in the GAN loss.

\subsection{Task Classifier}
\label{ssec:Sentiment}
Assuming the source-to-target generation ($G_{s\rightarrow t}$) does not change the sentiment label, we can train a transferable sentiment classifier over the generated intermediate domain: $f_t: G_{s\rightarrow t}(\mathcal{Z}) \rightarrow \mathcal{Y}$ using labels from the source domains:
\begin{equation}
    \mathcal{L}_{task} = -\mathbb{E}_{(\mathbf{z}, y)\sim (\mathcal{Z}_S, \mathcal{Y}_S)}\Big[-\log P\Big(y|f_t\big(G_{s\rightarrow t}(\mathbf{z})\big)\Big)\Big]
\end{equation}
After training, the classifier $f_t$ can be directly used in the target domain. To promote sentiment consistency between the generated intermediate representations and their original examples, we further backpropagate the task loss to the source-to-target generator.

\subsection{Joint Learning}
\label{ssec:Learning}
Our final objective is a weighted combination of different losses in the C-CycleGAN framework.\footnote{We omit the weights of each loss term for notation ease.}
For the model-based curriculum:
\begin{equation}
    \mathcal{L}_{\text{c-cyclegan}} = \mathcal{L}_{\text{cgan}}^{s\rightarrow t} + \mathcal{L}_{\text{cgan}}^{t\rightarrow s} + \mathcal{L}_{\text{cyc}} + \mathcal{L}_{\text{uni}}^t + \mathcal{L}_{\text{uni}}^s + \mathcal{L}_{\text{task}}
\end{equation}

For the model-free curriculum:
\begin{equation}
    \mathcal{L}_{\text{c-cyclegan}} = \mathcal{L}_{\text{cgan}}^{s\rightarrow t} + \mathcal{L}_{\text{cgan}}^{t\rightarrow s} + \mathcal{L}_{\text{cyc}} + \mathcal{L}_{\text{task}}
\end{equation}

This objective can be optimized by solving the following min-max game:
\begin{equation}
    f_t^* = \argmin_{f_t}\min_{\substack{G_{s\rightarrow t}\\G_{t\rightarrow s}\\h_t, h_s}}\max_{D_s, D_t}\mathcal{L}_{\text{c-cyclegan}}
\end{equation}

\section{Experiments}
In this section, we introduce the experimental settings and present results as well as analysis. Our source code will be released. 

\begin{table*}
\centering
\caption{Comparison with the state-of-the-art DA methods on Reviews-5 dataset. All numbers are percentages. The best class-wise and average classification accuracies trained on the source domains are emphasized in bold (similar below).}
\begin{tabular}{c|c|cccccc}
\toprule
Standards & Models & Camera & Laptop & Restaurant & Movie1 & Movie2 & Avg     \\ \midrule
\multirow{2}{*}{Source-only} & Single-best  & 68.8  & 62.5  & 64.0 & 76.9 & 75.8 & 69.6 \\
 & Source-combined & 69.6   & 71.5  & 68.5      & 77.0 & 76.7        & 72.7  \\ \midrule
\multirow{3}{*}{Single-best DA} & DANN~\cite{ganin2016domain} & 69.2   & 72.6  & 68.5 & 78.3 & 80.7 & 73.9  \\
 & ADDA~\cite{tzeng2017adversarial} & 69.4   & 73.2  & 69.6      & 79.1       & 81.5        & 74.6  \\
 & DAAN~\cite{yu2019transfer} & 69.4   & 73.8  & 71.6      & 79.5       & 82.8        & 75.4  \\ \midrule
\multirow{3}{*}{Source-combined DA} & DANN~\cite{ganin2016domain} & 67.0  & 73.3 & 68.2 & 77.4 & 80.8 & 73.3 \\
 & ADDA~\cite{tzeng2017adversarial} & 69.6   & 74.1  & 69.5      & 80.5       & 82.6        & 75.3  \\
 & DAAN~\cite{yu2019transfer} & 69.4   & 74.6  & 72.4      & 80.2       & 83.2        & 76.0  \\ \midrule
\multirow{4}{*}{Multi-source DA} & Autoencoder+MDAN~\cite{zhao2018adversarial} & 65.0 & 59.0   & 64.5 & 60.8 & 52.1 & 60.3  \\
 & MDAN (TextCNN)~\cite{zhao2018adversarial} & 68.0    & 72.0   & 71.0 & 77.4 & 78.7 & 73.4  \\
 &  CMSS~\cite{yang2020curriculum} & 71.8 & 75.4   & 73.3 & 81.2 & 85.6 & 77.5  \\
 & \textbf{C-CycleGAN (Ours)} & \textbf{73.8}   & \textbf{76.0}   & \textbf{76.0} & \textbf{82.0} & \textbf{87.5} & \textbf{79.1}  \\ \midrule
 Oracle& TextCNN & 76.8   & 77.5  & 77.5 & 84.4 & 90.6 & 81.4  \\
\bottomrule
\end{tabular}
\label{tab:five_domains_baseline}
\end{table*}

\begin{table*}
\centering
\caption{Comparison with the state-of-the-art DA methods on Reviews-5 dataset using BERT embedding.}
\begin{tabular}{c|c|cccccc}
\toprule
Standards & Models & Camera & Laptop & Restaurant & Movie1 & Movie2 & Avg     \\ \midrule
\multirow{2}{*}{Source-only} & Single-best  & 72.3  & 74.5  & 75.4 & 79.4 & 83.1 & 76.9 \\
 & Source-combined & 73.6   & 74.8  & 76.8      & 80.1 & 85.7        & 78.2  \\ \midrule
\multirow{1}{*}{Multi-source DA} & \textbf{C-CycleGAN (Ours)} & \textbf{76.9}   & \textbf{78.4}   & \textbf{79.7} & \textbf{83.1} & \textbf{88.3} & \textbf{81.3}  \\ \midrule
 Oracle& BERT &  78.3   &  79.5  &  81.2 & 85.1  & 90.8  & 83.0  \\
\bottomrule
\end{tabular}
\label{tab:five_domains_baseline_bert}
\end{table*}

\subsection{Experimental Settings}

\subsubsection{Datasets.}

We evaluate our approach using two combined datasets of cross-domain sentiment analysis: Reviews-5~\cite{yu2016learning} and Amazon benchmark~\cite{chen2012marginalized}. Each dataset contains multiple domains. For each dataset, we create multiple MDA settings by taking each domain as \emph{target}, and the rest as \emph{sources}.
In addition, we further consider a cross-lingual transfer setting using the Multilingual Amazon Reviews Corpus~\cite{chen2019multi}, to validate the generalizability of our approach to a broader family of transfer learning.

The \textbf{Reviews-5 dataset}~\cite{yu2016learning} includes five domains of customer reviews. \emph{Movie1} \cite{pang2005seeing} and \emph{Movie2}~\cite{socher2013recursive}
are movie reviews; \emph{Camera}~\cite{hu2004mining}
contains reviews of digital products such as MP3 players and cameras; \emph{Laptop} and \emph{Restaurant} are laptop and restaurant reviews respectively taken from SemEval 2015 Task 12~\cite{yu2016learning}. The training set sizes are 3,270, 1,707, 1,372, 9,162, and 8,113 for Movie1, Movie2, Camera, Laptop and Restaurant, respectively. The test size is 200 for all domains.

The \textbf{Amazon benchmark dataset}~\cite{chen2012marginalized} contains four domains of product reviews on Amazon: \emph{Books}, \emph{DVD}, \emph{Kitchen}, and \emph{Electronics}, with training set sizes of 6,465, 5,586, 7,681, and 7,945 respectively. The test size is 200 for all domains. This dataset has been preprocessed by the authors into TF-IDF representations, using the 5,000 most frequent unigram and
bigram features. Therefore, word order information is unavailable.

The \textbf{Multilingual Amazon Reviews Corpus}~\cite{chen2019multi} is a collection of Amazon reviews from four languages: \emph{German}, \emph{English}, \emph{French}, and \emph{Japanese}. For each language, there are three domains including Books, DVD, and Music. The training set size and test set size for each domain of each language are 52,000 and 2,000.

\subsubsection{Evaluation Metrics.} Following~\cite{zhao2018adversarial,guo2018multi}, we use classification accuracy as metric to evaluate the sentiment classification results. Larger values represent better performances.

\subsubsection{Baselines.}
We consider the following baselines:
\begin{enumerate}
    \item \textbf{Source-only}, directly training on the source domains and testing on the target domain, which includes two settings: single-best, the best test accuracy on target among all source domains;
    source-combined, the target accuracy of the model trained on the combined source domain.
    \item \textbf{Single-source domain adaptation methods}, including DANN \cite{ganin2016domain}, ADDA \cite{tzeng2017adversarial}, and DAAN \cite{yu2019transfer}, trained with both single-best and source-combined settings.
    \item \textbf{Multi-source domain adaptation models}, including state-of-the-art approaches MDAN \cite{zhao2018adversarial}, MoE \cite{guo2018multi}, and CMSS~\cite{yang2020curriculum}.
\end{enumerate}
We also report the results of an oracle setting, where the model is both trained and tested on the target domain.




\begin{table*}
\centering 
\caption{Comparison with the state-of-the-art DA methods on Amazon Benchmark dataset.}
\begin{tabular}{c|c|ccccc}
\toprule
Standards & Models & Books  & DVD   & Kitchen & Electronics & Avg  \\ \midrule
\multirow{2}{*}{Source-only} & Single-best & 75.4  & 81.3  & 86.5   & 86.5 & 82.4  \\
 & Source-combined & 76.5 & 81.6  & 86.7  & 85.3 & 82.5 \\ \midrule
\multirow{3}{*}{Single-best DA} & DANN~\cite{ganin2016domain} & 76.5  & 77.2  & 83.6   & 84.3 & 80.4    \\
 & ADDA~\cite{tzeng2017adversarial} & 74.4  & 78.2  & 82.6   & 82.1       & 79.3  \\
 & DAAN~\cite{yu2019transfer} & 77.2  & 76.8  & 83.5   & 86.5       & 81.0     \\ \midrule
\multirow{3}{*}{Source-combined DA} & DANN~\cite{ganin2016domain} & 77.9  & 78.9  & 84.9   & 86.4 & 82.0  \\
 & ADDA~\cite{tzeng2017adversarial} & 76.6  & 77.1  & 82.5   & 82.5       & 79.7  \\
 & DAAN~\cite{yu2019transfer} & 78.4  & 77.6  & 85.4   & 87.2       & 82.2   \\ \midrule
\multirow{4}{*}{Multi-source DA} & MDAN~\cite{zhao2018adversarial} & 78.0   & 85  & 85.3   & 86.3       & 82.5  \\
 & MoE~\cite{guo2018multi} & 78.9  & 81.3 & 87.4  & 87.9 & 83.9   \\
 &  CMSS~\cite{yang2020curriculum} & 78.1 & 80.2   & 87.2 & 87.2 & 83.2 \\
 & \textbf{C-CycleGAN (Ours)} & \textbf{80.3}  & \textbf{82.2}  & \textbf{88.9}   & \textbf{89.1} & \textbf{85.1} \\ \midrule
Oracle & TextCNN & 76.7 & 81.3 & 87.1 & 85.2 & 82.6 \\
\bottomrule
\end{tabular}
\label{tab:amazon_baseline}
\end{table*}

\subsubsection{Implementation Details.}
For the pre-training of text encoder, we use a 2-layer Bidirectional LSTM as \emph{encoder} and a 1-layer LSTM as \emph{decoder}. The initial learning rate is 0.00001 with a decay rate of 0.5 every 200 steps. The dimension of word embeddings and hidden states are both set to 256.
For experiments with BERT, we use the 12-layer ``bert-base-uncased'' version due to memory constraints.
The weights for $\mathcal{L}_{\text{cgan}}$, $\mathcal{L}_{\text{cyc}}$, $\mathcal{L}_{\text{uni}}$, and $\mathcal{L}_{\text{task}}$ are 0.1, 1, 1 and 1, respectively.
During decoding, we choose as input between the true previous token and the generated token with a probability of 0.5 of selecting either one.
For the Amazon benchmark dataset, we use the original TF-IDF feature vectors as the representation, without further encoding or pre-training.
We leverage a 4-layer multi-layer perceptron (MLP) to implement the generator and discriminator of CycleGAN, as well as the sentiment classifier.
The initial learning rate is 0.0001 with a decay rate of 0.5 every 100 steps. We use Adam~\cite{kingma2015adam} as the optimizer with beta1 of 0.5, beta2 of 0.999, batch size of 64, and weight decay of 0.0001.
In the multilingual transfer experiments, we obtain cross-lingual word embeddings by projecting the pre-trained monolingual word embeddings~\cite{bojanowski2017enriching} of the 4 languages into English (pivot language) using an unsupervised method~\cite{artetxe2017acl}.



\begin{table*}
\centering 
\caption{Comparison with the state-of-the-art DA methods on Multilingual Amazon Reviews Corpus dataset.}
\begin{tabular}{c | c | c c c c | c c c c}
\toprule
 \multirow{2}[0]{*}{Standards} & \multirow{2}[0]{*}{Models}& \multicolumn{4}{c}{German} & \multicolumn{4}{c}{English}  \\
 & & Books & DVD   & Music & Avg   & Books & DVD   & Music & Avg  \\\midrule
    \multirow{2}{*}{Source-only} &Single-best &63.6 &	64.7 &	64.9 &	64.4 &	65.3 &	62.5 &	63.3 &	63.7 \\
    & Source-combined& 61.5 &	64.6 &	63.6 &	63.2 &	63.7 &	65.0 &	60.1 &	63.0  \\ \midrule
    Multi-source DA &\textbf{C-CycleGAN (Ours)}& \textbf{78.3} &	\textbf{78.4} &	\textbf{79.1} &	\textbf{78.6} &	\textbf{78.0} &	\textbf{77.8} &	\textbf{79.0} &	\textbf{78.3}  \\ \midrule
    Oracle & TextCNN & 83.2 &	89.0 &	88.2 &	86.8 &	85.2 &	85.5 &	81.1 &	83.9  \\ \midrule \midrule
     \multirow{2}[0]{*}{Standards} & \multirow{2}[0]{*}{Models} & \multicolumn{4}{c}{French}    & \multicolumn{4}{c}{Japanese} \\
 & & Books & DVD   & Music & Avg   & Books & DVD   & Music & Avg \\\midrule
     \multirow{2}{*}{Source-only} &Single-best &	65.3 &	64.3 &	64.2 &	64.6 &	63.5 &	63.5 &	64.8 &	64.0 \\
    & Source-combined &	63.6 &	63.0 &	63.4 &	63.3 &	63.7 &	62.7 &	64.0 &	63.4 \\ \midrule
    Multi-source DA &\textbf{C-CycleGAN (Ours)}  &	\textbf{78.6} &	\textbf{77.6} &	\textbf{76.9} &	\textbf{77.7} &	\textbf{75.2} &	\textbf{74.9} &	\textbf{76.8} &	\textbf{76.2} \\ \midrule
    Oracle  & TextCNN & 88.3 &	77.6 &	84.1 &	83.3 &	60.4 &	61.8 &	69.4 &	69.4 \\
\bottomrule
\end{tabular}
\label{tab:cross_lingual_baseline_1}
\end{table*}

\subsection{Results on Reviews-5 Dataset}
\label{ssec:single-lingual}

We first evaluate our approach on the dataset of plain textual input: Reviews-5. We perform experiments with each domain as the target and the rest as sources.
Table~\ref{tab:five_domains_baseline} shows the performance of different DA methods and Table~\ref{tab:five_domains_baseline_bert} shows the extended results using BERT embedding~\cite{devlin2019bert}.
We have the following observations\footnote{The first 5 points are based on Table~\ref{tab:five_domains_baseline}, and the last point is based on Table~\ref{tab:five_domains_baseline_bert}.}:

(1) Without considering domain shift, both source-only settings, \textit{i.e.} single-best and source-combined, obtain poor accuracy: 69.6\% and 72.7\%, around 10\% worse than the oracle~(81.4\%). This motivates the research on domain adaptation.

(2) When directly applying to the MDA task, the single-source DA methods outperform the source-only setting.
Since customers' reviews vary a lot across domains, features related to sentiment also vary a lot. Therefore these DA methods that can make the domain gap smaller achieve better results than source-only setting.

\begin{table*}
\centering 
\caption{Ablation study on different components of the proposed C-CycleGAN framework on the Reviews-5 dataset.}
\begin{tabular}{l | cccccc}
\toprule
Models & \multicolumn{1}{l}{Camera} & \multicolumn{1}{l}{Laptop} & \multicolumn{1}{l}{Restaurant} & \multicolumn{1}{l}{Movie1} & \multicolumn{1}{l}{Movie2} & \multicolumn{1}{l}{Avg} \\\midrule
    CycleGAN~\cite{zhu2017unpaired} & 68.7 &	75.4 &	71.6 &	82.5 &	86.7 &	77.0 \\
    MDAN~\cite{zhao2018adversarial} + CycleGAN~\cite{zhu2017unpaired} & 70.8 &	75.2 &	71.2 &	79.9 &	86.2 &	76.7 \\
    \midrule
    CycleGAN~\cite{zhu2017unpaired}+CMSS~\cite{yang2020curriculum}  & 71.5 &	75.4 &	70.8 &	81.1 &	86.1 &	77.0 \\
    C-CycleGAN (model-based) & 72.8 &	75.7	 & 73.5 &	81.7 &	87.3 &	78.2 \\
    C-CycleGAN (model-free) & \textbf{73.8} &	\textbf{76.0} &	\textbf{76.0} &	\textbf{82.0} &	\textbf{87.5} &	\textbf{79.1} \\
\bottomrule
\end{tabular}
\label{tab:Ablation}
\end{table*}

(3) Comparing the performances of source-combined and single-best DA settings, we can find that sometimes naively performing single-source domain adaptation approaches on a combined dataset of different sources could produce worse result~(\textit{i.e.} 73.3\% of DANN) than on a single source domain~(\textit{i.e.} 73.9\% of DANN). This naturally motivates research on multi-source domain adaptation.

(4) Most of the state-of-the-art multi-source domain adaptation methods perform better than single-source domain adaptation methods by considering domain-invariant features and fusing information across all domains. However, MDAN~\cite{zhao2018adversarial}, which has been demonstrated to be effective on Amazon benchmark dataset, performs worse~(60.3\% and 73.4\%) than single-best DA settings~(\textit{e.g.} 74.6\% and 75.4\%). This indicates that some of the previous multi-source domain adaptation methods may be only effective on a certain kind of data representation (\textit{e.g.} bag-of-words or TF-IDF).

(5) C-CycleGAN performs the best (79.1\%) among all adaptation settings. Compared to the best results inside the Source-only, Single-best DA, Source-dombined DA and other Multi-source DA methods, C-CycleGAN achieves 6.4\%, 3.7\%, 3.1\% and 1.6\% performance boost, respectively.
These results demonstrate that the proposed C-CycleGAN model can achieve significant better performance compared to state-of-the-art methods. The performance improvements benefit from the advantages of C-CycleGAN.
First, an intermediate representation domain is generated with cycle-consistency and sentiment consistency which is closer to the target domain and preserves the annotation information of the source samples.
Second, the proposed weighting mechanisms can dynamically assign weights to different source samples, which takes into account the source samples' similarity to the target and enhances the adaptation performance.
Finally, the text reconstruction in the pre-trained text encoder minimizes the information loss during the feature encoding process.

(6) BERT embedding performs much better than Bi-LSTM for all the methods, which demonstrates the superiority of BERT in learning pre-trained embeddings. The proposed C-CycleGAN achieves 3.1\% performance gains as compared to the best source-only setting.

\subsection{Results on Amazon Benchmark Dataset}
\label{ssec:single-lingual2}

Table~\ref{tab:amazon_baseline} shows the results on the Amazon benchmark dataset, which takes TF-IDF as text representations. 
We can observe that:

(1) Comparing the performance of source-only~(82.5\%) and Oracle~(82.6\%), we can see that the domain gap between sources and target is less than 1\%, much smaller than the domain gap of Reviews-5~(>10\%). This indicates that the data representation type of the datasets is closely associated with how large the domain gap is.

(2) Several multi-source adaptation methods~(\textit{e.g.} MoE~\cite{guo2018multi}) perform even better than Oracle. This is because that the domain gap is relatively small and multi-source adaptation leverages more information from multiple domains than Oracle, which only has access to the samples from the target. This further indicates the importance of diverse data from different source domains.

(3) The proposed C-CycleGAN has the best performance (85.1\%) among all approaches with 1.2\% and 2.5\% better classification accuracy than MoE and Oracle respectively. Compared to other methods (\textit{e.g.} MDAN) whose performance fluctuates significantly across datasets (Reviews-5 and Amazon Benchmark datasets), the proposed C-CycleGAN can provide consistent superior performance across datasets.

\subsection{Multilingual Transfer Experiments}
\label{ssec:multi-lingual}

We also perform experiments on the Multilingual Amazon Reviews Corpus. For each category domain (Books, DVD, Music) of each language, we perform adaptation to it with datesets of the same category domain from other languages as sources. Table~\ref{tab:cross_lingual_baseline_1} shows the performance of different adaptation methods.
We can observe that:

(1) The proposed C-CycleGAN achieves the best performance of all DA methods across all languages and on all category domains.

(2) In most cases, Oracle gives the best performance; however, in several settings, C-CycleGAN can achieve similar or even better results than the oracle (\textit{e.g.} 77.6\% and 77.6\% for DVD in French; 76.8\% and 69.4\% for Music in Japanese). This further demonstrate that our framework has a wide range of applicability, not only across different types of data representation, but also across different languages.

\begin{table*}
\centering 
\caption{Ablation study on the influence of cycle-consistency in C-CycleGAN on the Reviews-5 dataset.}
\begin{tabular}{l | cccccc}
\toprule
Models & \multicolumn{1}{l}{Camera} & \multicolumn{1}{l}{Laptop} & \multicolumn{1}{l}{Restaurant} & \multicolumn{1}{l}{Movie1} & \multicolumn{1}{l}{Movie2} & \multicolumn{1}{l}{Avg} \\\midrule
   C-CycleGAN w/o cycle-consistency & 72.2 &	73.1 &	72.0 &	80.3 &	85.1 &	76.5 \\
    C-CycleGAN w cycle-consistency & \textbf{73.8} &	\textbf{76.0} &	\textbf{76.0} &	\textbf{82.0} &	\textbf{87.5} &	\textbf{79.1} \\
\bottomrule
\end{tabular}
\label{tab:Ablation_CycleConsistency}
\end{table*}

\subsection{Ablation Study}

We conduct a series of ablation studies on the Reviews-5 dataset to demonstrate the improvements of C-CycleGAN over existing state-of-the-art approaches.
The results are described in Table~\ref{tab:Ablation}, where all CycleGANs are performed in a source-combined manner.

First, we investigate whether it is necessary to align the representations before applying CycleGAN. ``MDAN + CycleGAN'' in Table~\ref{tab:Ablation} represents first aligning the encoded representations using MDAN and then applying CycleGAN. Comparing the first two rows in Table~\ref{tab:Ablation}, we can see applying MDAN before CycleGAN achieves worse performance, which indicates that it is unnecessary to perform additional alignment before CycleGAN. This is probably because extracting the domain-invariant features between the source and target domains might lose some discriminative features in the target domain that are related to sentiment.

\begin{figure*}[t]
\centering
\includegraphics[width=0.94\linewidth]{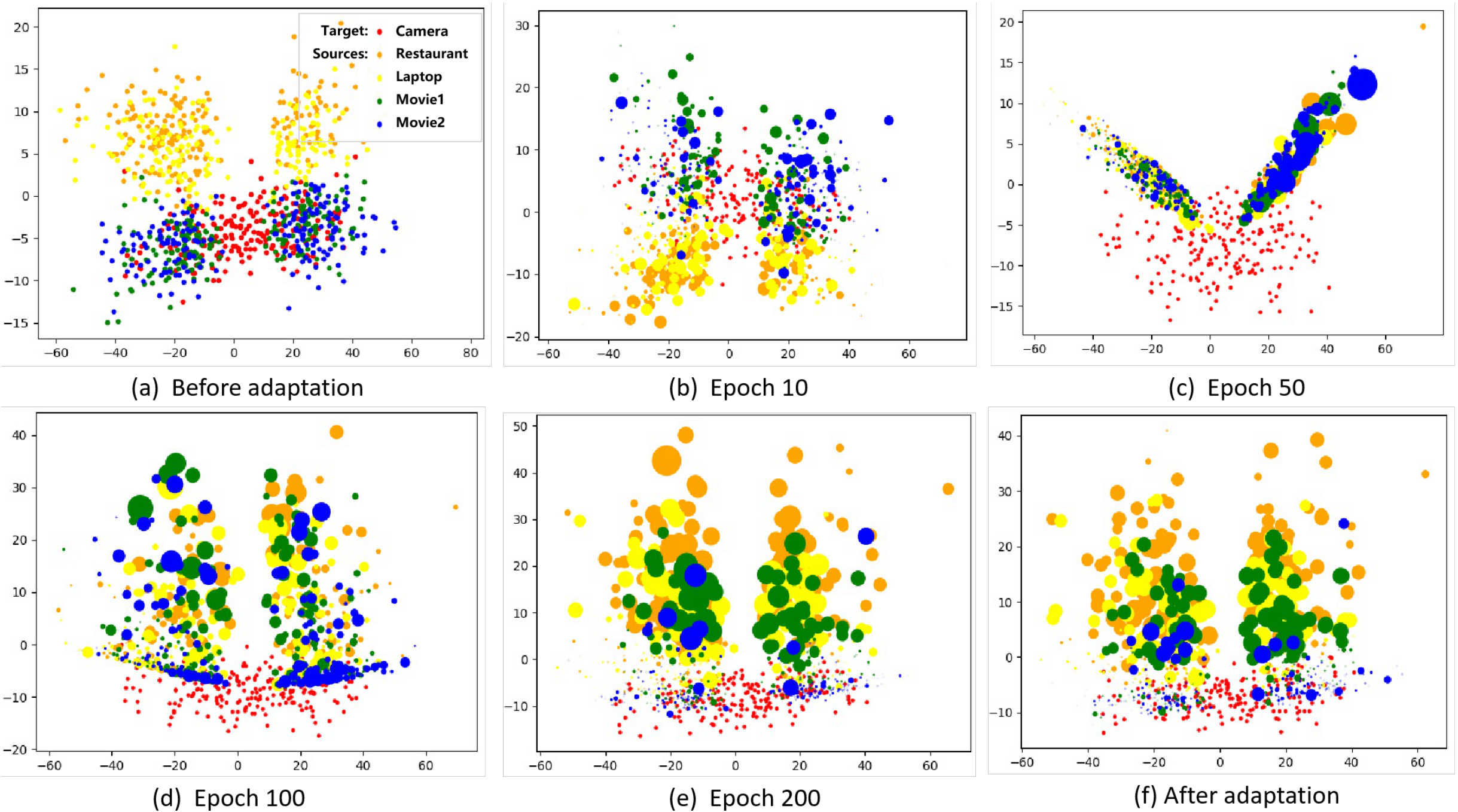}
\caption{Visualization of feature spaces in different training stages of C-CycleGAN on the Reviews-5 dataset. Target samples are in red, while source samples are in other colors. Point size denotes the similarity of each source sample to the target domain obtained from output of the domain discriminator. For better visualization, smaller points represent samples closer to the target domain.
}
\label{fig:dynamic_shape}
\end{figure*}

Second, we investigate the effectiveness of the proposed model-based and model-free weighting methods. From the last three rows, we can see that compared to CMSS~\cite{yang2020curriculum}, the proposed model-based and model-free weighting schemes improve accuracy by 1.2\% and 2.1\% respectively. Because CMSS takes the original source samples as input to compute the weights, it cannot reflect the dynamic changing of source samples' weights. The proposed model-based weighting mechanism is based on the generated intermediate domain, which itself dynamically changes.
The model-based method requires an additional network to compute the similarity to the target domain, which not only increase the computation cost, but also takes longer to learn the discriminative patterns between sources and target, before which CycleGAN may learn the wrong patterns.

Finally, we evaluate the influence of cycle-consistency in the proposed C-CycleGAN model. As in~\cite{zhu2017unpaired}, we find that standard adversarial procedures without cycle-consistency often lead to the mode collapse problem, where all input representations are mapped to the same output representation and the optimization fails to make progress.
The comparison between with and without cycle-consistency in C-CycleGAN on the Reviews-5 dataset is shown in Table~\ref{tab:Ablation_CycleConsistency}. The result comparison (79.1 vs. 76.5) clearly demonstrates the effectiveness and necessity of cycle-consistency.

\subsection{Visualization}

In this section, we visualize the features of source and target samples during different training stages of C-CycleGAN.

By using PCA to reduce the dimensionality of samples, we project samples from five domains in Reviews-5~\cite{yu2016learning} onto a 2-dimensional plane in different stages of training. The visualization results are shown in Figure~\ref{fig:dynamic_shape}.
We can conclude that during the training process, all source domains get closer to the target domain. At the same time, we can see that the samples far from the target domain can be well differentiated by the discriminator, and are assigned with smaller weights (larger points).

Figure~\ref{fig:dynamic_shape} (a) and (f) visualize the representation space before and after adaptation correspondingly. We can see that the samples in Movie1 and Movie2 are the closest since they are all about reviews in movies. Movie1 is also closer with Camera and Laptop after adaptation, which is desirable because these domains involve common reviews on image quality or upgrade of electronics. For example, the Camera domain may have reviews like ``Picture is clear and easy to carry. Love SONY.''; while in Movie1: ``Transitions smoothly and the image quality is clean'', and in Laptop: ``The 4K display is so sharp, the slim book is so light in a bag''. We can hardly distinguish which domains these reviews belong to without prior information.

\begin{figure*}[t]
\centering
\includegraphics[width=1.0\linewidth]{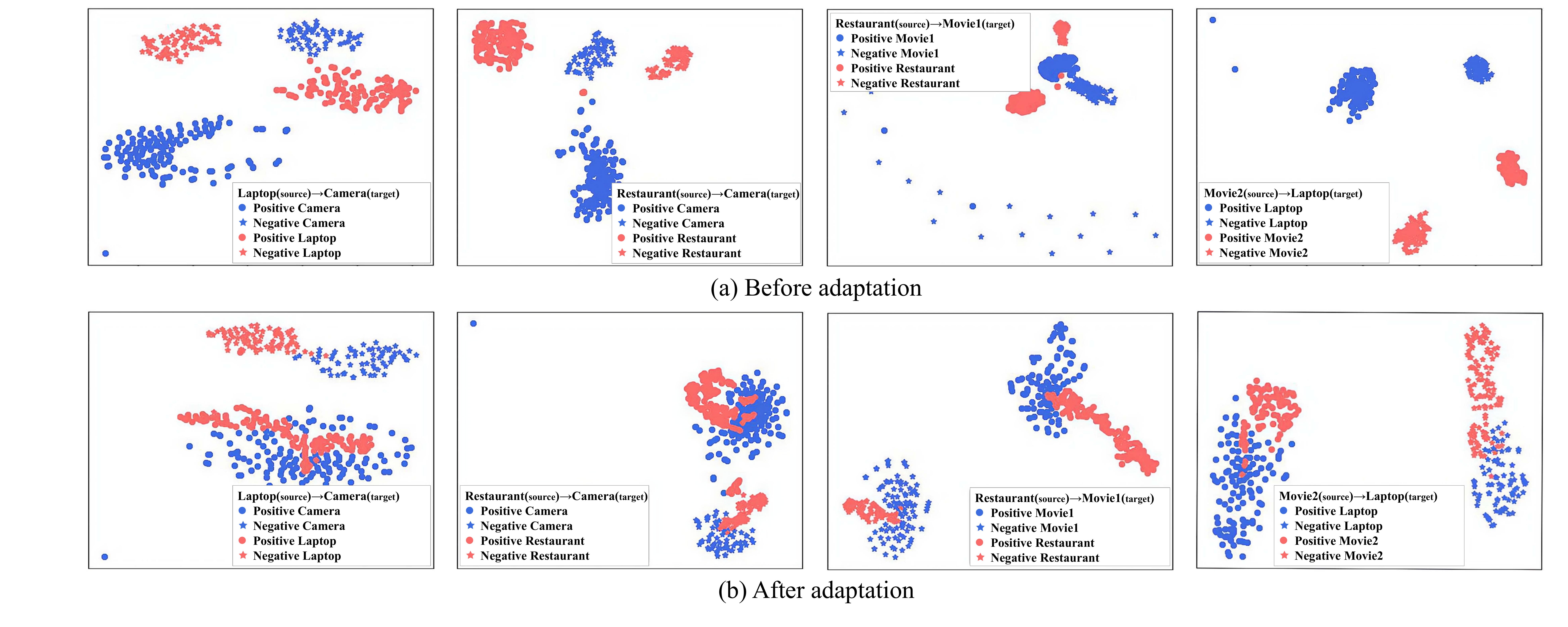}
\caption{t-SNE visualization of the features before and after adaptation on the Reviews-5 dataset. \textcolor{my_red}{Red} represents source features and \textcolor{my_blue}{Blue} represents target features.
}
\label{fig:five_domains_visual}
\end{figure*}



We further plot the learned features with t-SNE~\cite{maaten2008visualizing} on four adaptation settings, with the results shown in Figure~\ref{fig:five_domains_visual}. The top row represents the feature embeddings before adaptation, while the bottom row represents the feature embeddings after adaptation by C-CycleGAN. Red represents source features and Blue represents target features. As we can see, before adaptation, the source samples can be obviously classified but such classifier cannot work well on the target samples; with the proposed C-CycleGAN, source and target features of the same class become more aggregated after adaptation. These observations further demonstrate the effectiveness of C-CycleGAN. 


\section{Conclusion}
In this paper, we proposed a novel multi-source domain adaptation framework, named curriculum cycle-consistent generative adversarial network (C-CycleGAN), for textual sentiment classification. C-CycleGAN contains three main component: pre-trained text encoder for encoding text instances into a latent continuous representation space with minimal information loss; intermediate domain generator with curriculum instance-level adaptation considering the importance of different source samples; and task classifier to perform the final sentiment classification.
The generated intermediate domain bridges the domain gap between the source and target domains, while preserving the sentiment semantics. The proposed dynamic model-based and model-free weighting mechanisms can assign higher weights to the source samples that are closer to the target domain. Further, C-CycleGAN does not require prior domain labels of source samples, which makes it more practical in real-world scenarios. Extensive experiments on multiple benchmark datasets demonstrate that C-CycleGAN significantly outperforms existing state-of-the-art DA methods. In future studies, we plan to construct a large-scale textual dataset with more fine-grained sentiment categories and extend our framework to corresponding MDA tasks. We will explore multi-modal domain adaptation by jointly modeling multiple modalities, such as image and text.

\begin{acks}
This work is supported by Berkeley DeepDrive, the Major Project for New Generation of AI Grant (No. 2018AAA0100403), the National Natural Science Foundation of China (Nos. 61701273, 61876094, U1933114), Natural Science Foundation of Tianjin, China (Nos. 20JCJQJC00020, 18JCYBJC15400, 18ZXZNGX00110), and the Fundamental Research Funds for the Central Universities.
\end{acks}

\bibliographystyle{ACM-Reference-Format}
\balance\bibliography{main}


\end{document}